\documentclass[10pt,journal,compsoc]{IEEEtran}
\ifCLASSOPTIONcompsoc
  \usepackage[nocompress]{cite}
\else
  \usepackage{cite}
\fi
\ifCLASSINFOpdf
\else
\fi
\usepackage{hyperref}
\usepackage{url}
\usepackage{wrapfig}
\usepackage{amsmath,amsfonts}
\usepackage{algorithmic}
\usepackage{algorithm}
\usepackage{array}
\usepackage[caption=false,font=normalsize,labelfont=sf,textfont=sf]{subfig}
\usepackage{textcomp}
\usepackage{stfloats}
\usepackage{url}
\usepackage{verbatim}
\usepackage[pdftex]{graphicx}
\usepackage{xcolor}
\usepackage{bm}
\usepackage{bbding}
\usepackage{pifont}
\usepackage{makecell}
\usepackage[T1]{fontenc}
\usepackage[utf8]{inputenc}
\usepackage{booktabs}       
\usepackage{nicefrac}       
\usepackage{microtype}      
\usepackage{epstopdf}
\usepackage{multirow}
\usepackage{mathtools}
\usepackage{amsmath}
\DeclareMathOperator*{\argmax}{arg\,max}
\usepackage[dvipsnames]{xcolor}

\begin{document}
%
\title{Point-Supervised Facial Expression Spotting with Gaussian-Based Instance-Adaptive Intensity Modeling}
%
%
%
%

\author{Yicheng~Deng,
Hideaki~Hayashi,~\IEEEmembership{Member,~IEEE,}
and~Hajime~Nagahara,~\IEEEmembership{Member,~IEEE}

\IEEEcompsocitemizethanks{\IEEEcompsocthanksitem Y. Deng is with the Graduate School of Information Science and Technology, The University of Osaka, Suita, 565-0871, Japan. Email: yicheng@is.ids.osaka-u.ac.jp\protect\\
\IEEEcompsocthanksitem H. Hayashi and H. Nagahara are with the D3 Center, The University of Osaka, Suita, 565-0871, Japan. Email: hayashi@ids.osaka-u.ac.jp; nagahara@ids.osaka-u.ac.jp}
}

%
%

\markboth{Journal of \LaTeX\ Class Files,~Vol.~14, No.~8, August~2015}%
{Shell \MakeLowercase{\textit{et al.}}: Bare Demo of IEEEtran.cls for Computer Society Journals}
%



\IEEEtitleabstractindextext{%
\begin{abstract}
Automatic facial expression spotting, which aims to identify facial expression instances in untrimmed videos, is crucial for facial expression analysis. Existing methods primarily focus on fully-supervised learning and rely on costly, time-consuming temporal boundary annotations. In this paper, we investigate point-supervised facial expression spotting (P-FES), where only a single timestamp annotation per instance is required for training. 
We propose a unique two-branch framework for P-FES. First, to mitigate the limitation of hard pseudo-labeling, which often confuses neutral and expression frames with various intensities, we propose a Gaussian-based instance-adaptive intensity modeling (GIM) module to model instance-level expression intensity distribution for soft pseudo-labeling. By detecting the pseudo-apex frame around each point label, estimating the duration, and constructing an instance-level Gaussian distribution, GIM assigns soft pseudo-labels to expression frames for more reliable intensity supervision. The GIM module is incorporated into our framework to optimize the class-agnostic expression intensity branch. Second, we design a class-aware apex classification branch that distinguishes macro- and micro-expressions solely based on their pseudo-apex frames. 
During inference, the two branches work independently: the class-agnostic expression intensity branch generates expression proposals, while the class-aware apex-classification branch is responsible for macro- and micro-expression classification.
Furthermore, we introduce an intensity-aware contrastive loss to enhance discriminative feature learning and suppress neutral noise by contrasting neutral frames with expression frames with various intensities. Extensive experiments on the SAMM-LV, CAS(ME)$^2$, and CAS(ME)$^3$ datasets demonstrate the effectiveness of our proposed framework. Code is available at \url{https://github.com/KinopioIsAllIn/GIM}.
\end{abstract}

}

\maketitle

\IEEEdisplaynontitleabstractindextext

%
\IEEEpeerreviewmaketitle

\IEEEraisesectionheading{\section{Introduction}\label{sec:introduction}}

%
%
%
%
\IEEEPARstart{F}{acial} expressions serve as a crucial means of nonverbal communication, playing a key role in conveying human emotions. They can be broadly categorized into macro-expressions (MaEs) and micro-expressions (MEs). MaEs are typically high-intensity, involve the global facial area, and persist for 0.5 to 4.0 seconds~\cite{ekman2003darwin}. MaE analysis is valuable in various fields, including social robotics~\cite{rawal2022facial} and virtual reality~\cite{ortmann2023facial}. In contrast, MEs are subtle \cite{li2025could}, rapid (lasting less than 0.5 seconds) \cite{ben2021video} and often appear in the local facial area. Despite being difficult to detect, they are important in emotion-related applications, such as lie detection~\cite{ekman1969nonverbal} and psychological counseling~\cite{nummenmaaekman}, as they are involuntary and reveal real emotions. Given their importance, both MaE and ME analysis contribute significantly to understanding human emotions and behavior.

\begin{figure}[t]
\centering
\includegraphics[width=0.95\linewidth]{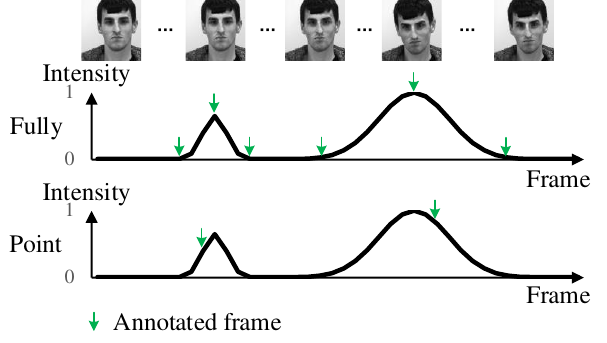}
\caption{Comparison of different forms of supervision. The fully-supervised method requires annotating the onset, apex, and offset frames for each instance, whereas the point-supervised method only requires annotating a single frame for each instance.}
\label{PSIntro}
\end{figure}

A requisite for efficient facial expression analysis is the detection of these expression clips in untrimmed videos, which is known as facial expression spotting (FES). As a preliminary step to recognizing the specific emotional types of facial expressions, FES focuses on localizing facial expression instances in untrimmed videos, determining their onset and offset frames while classifying each instance as either a MaE or an ME. FES is essential for accurately detecting expressions in videos, thus improving emotion recognition and advancing applications in human-computer interaction. 

Prior studies~\cite{yin2023aware, yu2023lgsnet, deng2024multi} have mainly focused on fully-supervised FES (F-FES). These approaches typically extract optical flow features. With the help of frame-level annotations, they either estimate the probability of expression occurrence at each temporal point or regress offsets toward the expression boundaries. While such precise supervision has driven substantial advances in F-FES, obtaining frame-level annotations is labor-intensive and limits practical applications, motivating the exploration of semi-supervised problem settings.

\begin{figure*}[tbp]
\centering
\includegraphics[width=0.95\textwidth]{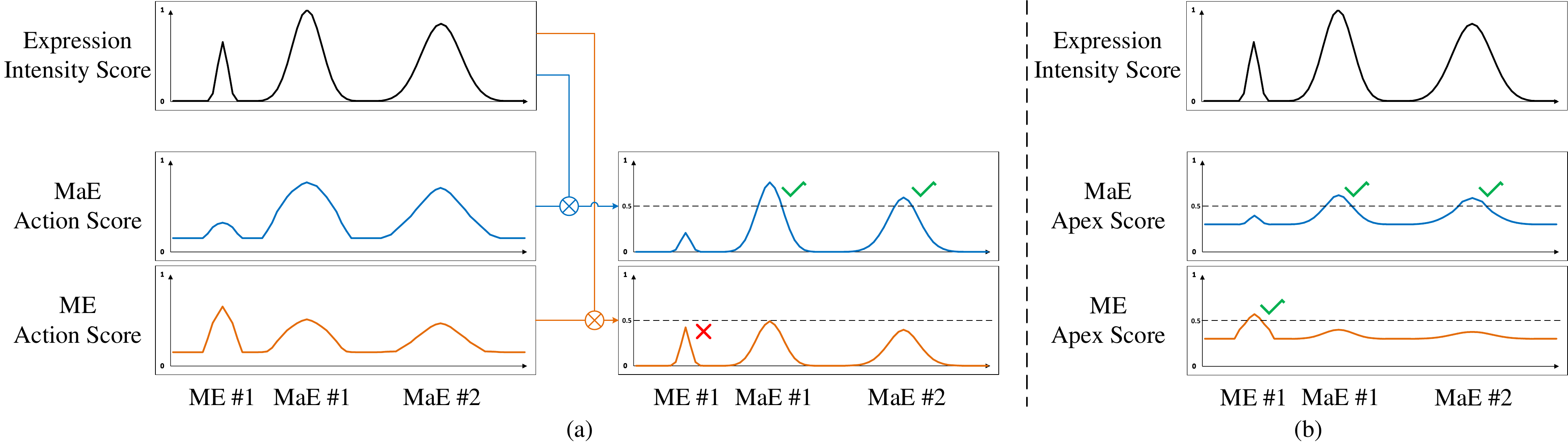}
\caption{Motivation illustration of the re-designed overall framework.
(a) General two-branch frameworks fuse the action scores with the expression intensity scores during both training and inference. However, since MEs typically exhibit lower intensity than MaEs, the fused ME scores are often suppressed, making MEs harder to spot.
(b) In our re-designed framework, both branches are optimized independently, preventing MEs from being overshadowed during inference.}
\label{two_opt_diff}
\end{figure*}

\begin{figure}[t]
\centering
\includegraphics[width=0.95\linewidth]{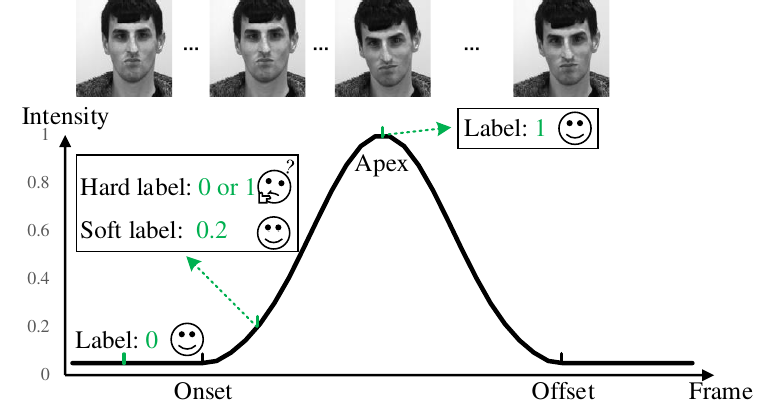}
\caption{Motivation illustration of the soft pseudo-labeling. Due to the fact that expression frames have various intensities, it is difficult to describe this characteristic by hard pseudo-labeling. We use soft pseudo-labeling to learn the intensity distribution of each instance, reducing the ambiguity in distinguishing between neutral and expression frames with various intensities.}
\label{abs}
\end{figure}

To adapt the insights from F-FES studies to more practical scenarios with limited annotation costs, this paper investigates point-supervised FES (P-FES). As illustrated in Fig.~\ref{PSIntro}, in contrast to F-FES, which requires annotating the onset and offset frames with low expression intensity, P-FES requires only a single timestamp annotation at any intensity for each instance. This approach can significantly reduce the annotation burden and time required for training models, making it more feasible to deploy in real-world applications. The potential challenge in P-FES lies in detecting subtle MEs while suppressing the effects of neutral movements, such as meaningless head movements or eye blinking. Unfortunately, little research has been conducted to address these challenges, leaving this task largely unexplored.

Despite the notable lack of research on P-FES, considerable efforts have been devoted to point-supervised temporal action localization (P-TAL), which shares the same problem setting as our task, with the only difference being the target domains. Previous P-TAL studies \cite{ma2020sf, lee2021learning, zhang2024hr} generally employ a two-branch framework for class-agnostic score estimation and action classification. Subsequently, they propose unique methods to assign hard pseudo-labels to reliable pseudo-action frames based on class-agnostic scores and feature similarity. 

However, we observe two limitations of directly applying current P-TAL framework to P-FES task. \textbf{First}, existing P-TAL frameworks typically fuse the outputs of two branches for action classification during training. Consequently, both outputs are also fused to generate action proposals during inference. While such a framework works well in the idealized setting of P-FES, we observe that it causes MEs to be overshadowed, as illustrated in Fig.~\ref{two_opt_diff} (a), since MEs usually exhibit lower intensity than MaEs. \textbf{Second}, the hard pseudo-labeling strategy may fail to help the model distinguish between neutral and expression frames with various intensities. As illustrated in Fig.~\ref{abs}, although the low-intensity expression frames are similar to the neutral frames in intensity, they should be assigned a label of 1 when using hard pseudo-labeling, just like the high-intensity frames near the apex frames. In this case, hard pseudo-labels cannot precisely describe the characteristics of expression intensity, resulting in inaccurate class-agnostic output scores.

To overcome the above limitations, we propose a unique two-branch framework specifically designed for P-FES, which consists of a regression-based class-agnostic expression intensity branch and a class-aware apex classification branch. The proposed framework contains three key technical features. 1) In this paper, we assume that the expression intensity in each instance follows an individual smooth Gaussian distribution. Based on this assumption, we propose Gaussian-based instance-adaptive Intensity Modeling (GIM) for P-FES. Specifically, for the class-agnostic expression intensity branch, we first detect the pseudo-apex frame around each labeled frame. Then, we estimate the rough duration for each expression instance based on the intensity scores. Subsequently, we build a Gaussian distribution for each expression instance, using the pseudo-apex frame and the feature distances to other pseudo-expression frames within the duration. Finally, we assign soft pseudo-labels as the expression intensity values for supervision and optimize the expression intensity branch. 2) For the class-aware apex classification branch, we focus on distinguishing between MaEs and MEs solely by their pseudo-apex frames, as the apex frame always contains the most important information about a facial expression instance. Unlike general two-branch frameworks, which optimize both branches by fusing their outputs for action classification during training, we optimize each branch separately. Therefore, during inference, two branches work independently to prevent MEs from being overshadowed. 3) We introduce an Intensity-Aware Contrastive (IAC) loss on reliable pseudo-labeled frames from different classes. This loss enhances the model’s ability to distinguish neutral frames from expression frames with various intensities, while suppressing neutral noise and emphasizing informative expression frames.

This paper is an extension of our previous conference publication \cite{deng2025gaussianbased}. The most significant updates is the redesigned overall framework. In the previous conference version, we employed a general two-branch framework following prior P-TAL works. In this paper, as shown in Fig.~\ref{two_opt_diff}(b), we redesign the framework such that the outputs of the two branches are no longer fused during either training or inference, and we introduce a class-aware apex classification branch to improve MaE and ME classification. We re-conducted all experiments, and the comparisons demonstrate superior performance over the previous conference version, particularly in ME spotting. Additionally, we present a more comprehensive qualitative and quantitative evaluation in the experimental section, including additional experiments on a new large-scale dataset CAS(ME)$^3$, an evaluation of apex frame detection, and more extensive ablation studies. The contributions are listed as follows, where * denotes the new contributions of this paper:
\begin{itemize}
    \item * We analyze the limitations of directly applying current P-TAL frameworks to P-FES and identify two main issues: (1) general two-branch frameworks fuse the outputs of both branches during training and inference, which consequently leads to MEs being overshadowed; and (2) hard pseudo-labeling makes it ambiguous to distinguish between neutral frames and expression frames with various intensities.
    \item * To address the first issue, we re-design the overall framework by separately processing the outputs of two branches for both training and inference. In addition, a class-aware apex classification branch is consequently designed specifically for P-FES, which focuses on distinguishing MaEs and MEs solely by their pseudo-apex frames, as the apex frame typically contains the most important information about a facial expression. 
    \item To address the second issue, we propose a Gaussian-based instance-adaptive Intensity Modeling module for soft pseudo-labeling. The GIM module adaptively models the intensity distribution of each expression proposal and assigns soft pseudo-labels to potential expression frames for supervision. 
    \item We introduce intensity-aware contrast learning on pseudo-labeled frames from different classes with various intensities, enhancing the discriminative feature learning and suppressing neutral noise.
    \item * We propose evaluating the accuracy of apex frame detection in the FES task. Even though apex frame detection is not required for the FES task, it is significant for further expression recognition.
    
\end{itemize}

\begin{figure*}[tbp]
\centering
\includegraphics[width=0.9\textwidth]{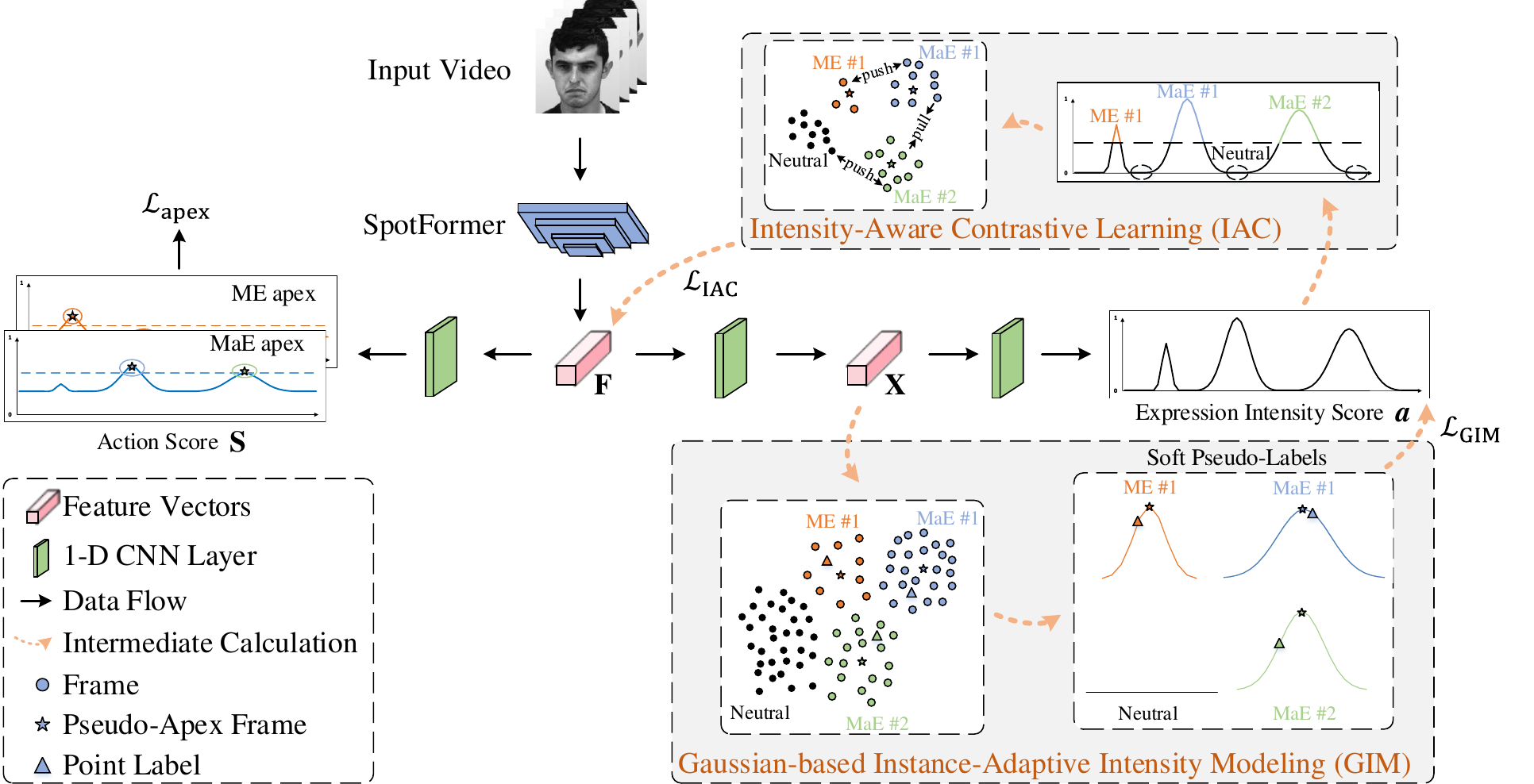}
\caption{Overview of the proposed framework. The framework initially calculates the optical flow and extracts snippet features by SpotFormer. These features are fed into a two-branch framework to obtain class-agnostic expression intensity scores (right) and class-aware apex scores (left). A GIM module is employed to build the Gaussian distribution for each expression instance and assign soft pseudo-labels to model the intensity distribution. An IAC module is employed to build contrasts among pseudo-labeled frames with various intensities to enhance feature learning and suppress neutral noise.}
\label{framework}
\end{figure*}

\section{Related Works}
\subsection{Facial expression spotting}
Previous FES methods can be grouped into traditional methods and deep learning methods. 
Traditional methods generally extract optical flow and analyze the pattern of each region of interest. He~\cite{yuhong2021research} proposed to eliminate the influence of head movement by the optical flow of the nose region. Zhao \textit{et al.}~\cite{zhao2022rethinking} refined feature extraction and employed a Bayesian optimization algorithm for analyzing optical flow patterns. Qin \textit{et al.}\cite{qin2023micro} employed a finer crop-align technique for facial alignment, followed by optical flow feature extraction and low-pass filtering to suppress high-frequency noise, thereby generating high-quality proposals. Yu \textit{et al.}\cite{yu2023efficient} further refined the processes of facial alignment, regions of interest extraction, and optical flow computation to enhance detection performance. Wang \textit{et al.}~\cite{wangunique} calculated skip-k-frame block-wise main directional mean optical flow (MDMO) \cite{liu2015main} features and analyzed the unique M-pattern of these features to detect MEs.

Recently, with advancements in deep learning techniques, many researchers have developed neural networks to tackle the F-FES task. Leng \textit{et al.}~\cite{leng2022abpn} proposed ABPN by extending BSN~\cite{lin2018bsn} which was originally designed for temporal action localization and adapting it for FES. Yin \textit{et al.} \cite{yin2023aware} refined ABPN~\cite{leng2022abpn} by introducing action unit-aware graph convolutional networks. Yu \textit{et al.}~\cite{yu2023lgsnet} designed a two-branch framework based on A2Net \cite{yang2020revisiting} and introduced additional attention modules for FES. Xu \textit{et al.}~\cite{xu2023integrating} proposed a framework that integrates VideoMAE-based feature extraction, multi-scale candidate segment generation, and a multi-start-point optical flow filtering method for accurate ME and MaE detection. Deng \textit{et al.}~\cite{deng2024multi} proposed an SW-MRO feature and introduced SpotFormer \cite{deng2024spotformer} to significantly improve the ME spotting performance.

\subsection{Point-supervised temporal action localization}
P-TAL shares the same problem setting as our task, differing only in the target domains, and was first introduced by Ma \textit{et al.}~\cite{ma2020sf}. Ju \textit{et al.}~\cite{ju2021divide} introduced a two-stage approach, which divides the video into multiple clips based on labeled frames and then performs regression and classification in each clip. Lee \textit{et al.}~\cite{lee2021learning} proposed to find the optimal sequence consistent with point labels based on the completeness score. Fu \textit{et al.}~\cite{fu2022compact} measured the confidence of each frame based on the feature similarity and rectified the output scores to assign reliable hard pseudo-labels. Zhang \textit{et al.}~\cite{zhang2024hr} proposed a two-stage framework comprising two reliability-aware stages that efficiently propagate high-confidence cues from point annotations at both the snippet and instance level. Xia \textit{et al.}~\cite{xia2024realigning} claimed that the most informative frame tends to appear in the central region of each instance. To this end, they introduced a plug-in proposal learning framework for realigning confidence with proposals.

Although P-FES, the focus of this study, has many similarities with P-TAL, research specifically dedicated to P-FES is extremely limited. Wang \textit{et al.}~\cite{wang2025micro} investigated a similar but unique micro-expression key frame inference task. Yu \textit{et al.}~\cite{yu2024weak} explored P-FES by employing a framework similar to general P-TAL methods.

In this paper, instead of employing a general P-TAL framework, we propose a new two-branch framework for expression intensity modeling and class-aware apex classification.


\subsection{Soft pseudo-labeling}
Soft pseudo-labeling is a widely used semi-supervised learning method, mainly explored in classification tasks. Unlike hard pseudo-labeling, which assigns the most confident class label to unlabeled data, this approach produces soft labels that reflect the full probability distribution, accounting for prediction uncertainty. Nassar \textit{et al.}~\cite{nassar2023protocon} introduced PROTOCON, which improves soft pseudo-labeling by utilizing information from neighboring samples in a prototypical embedding space, specifically for semi-supervised image classification. Similarly, Lukov \textit{et al.}~\cite{lukov2022teaching} proposed a method that refines logits by smoothing high-confidence classes based on their confidence levels while assigning a fixed low probability to low-confidence classes, effectively mitigating noisy labels in in-the-wild facial expression recognition.
Recently, several researchers \cite{liang2022gmmseg, wu2023sparsely, shen2024cgmgm} constructed class-wise Gaussian Mixture Models to model class-wise feature distribution for semantic segmentation.
Inspired by these works, we propose to construct an individual Gaussian distribution for each expression instance to assign soft pseudo-labels as direct intensity supervision signals. A regression model is then trained to learn the expression intensity distribution, instead of modeling the distribution of class probabilities. To the best of our knowledge, we are the first to investigate the application of soft pseudo-labeling for P-FES, providing a novel perspective on modeling expression intensity.

\section{Methodology}
\label{Methodology}
\subsection{Problem formulation}

Given an untrimmed facial video $\mathbf{V}=(\mathbf{v}_i)_{i=1}^{T}$ including a total of $T$ frames, we only annotate single timestamp for each facial expression instance and get the point set $\mathbf{Y}={(p_i, \bm{y}_i)}^N_{i=1}$, where $p_i$ represents the index of the annotated frame of the $i$-th expression instance during training, $N$ denotes the total number of ground-truth expression instances, and $\bm{y}_i$ denotes the multi-hot vector representing the action class (i.e., MaE and ME), respectively. Our goal is to detect as many expression instances as possible, which involves determining the boundary frames and classifying each instance as MaE or ME.

\subsection{Motivation}
As discussed in Section \ref{sec:introduction}, existing methods suffer from two main limitations. These two limitations motivate us to develop the framework with unique designs to address these issues through explicit intensity modeling and class-aware apex frame classification.

Recall that the first limitation of general P-TAL frameworks is that the outputs of the two branches are combined during both training and inference, which leads to degraded ME spotting performance. This coupling makes the framework less sensitive to ME spotting, suggesting the need for a decoupled framework and a more discriminative branch design. Therefore, we propose to decouple the two branches. Each branch is optimized independently during training, allowing their outputs to be processed separately at inference. Moreover, we revise the action classification branch to distinguish MaEs from MEs solely based on their pseudo-apex frames, as the optical flow feature at the apex frame usually provides the most critical information about a facial expression instance in terms of both intensity and duration~\cite{deng2024spotformer}. The re-designed overall framework is described in Section~\ref{owork}, the details of the new separate optimization strategy are introduced in Sections~\ref{sec:lossfunction}, and the inference process is introduced in Section~\ref{sec:inference}.

To overcome the second limitation that the hard pseudo-labeling strategy introduces ambiguity in FES when distinguishing between neutral and expression frames with various intensities, we propose the GIM module to assign soft pseudo-labels to frames with different intensity levels, thereby modeling the expression intensity distribution of each instance. Since facial expressions inherently evolve through the onset–apex–offset progression. Even MEs, which may appear as sudden and short-lived responses, follow a similar temporal evolution, making the proposed GIM module appropriate for modeling their intensity variations.
In addition, we convert the binary classification-based class-agnostic branch into a regression-based class-agnostic expression intensity branch to enable continuous estimation of expression intensity. The details are described in Section~\ref{gimsection}.

\subsection{Overall framework}
\label{owork}
Fig.~\ref{framework} shows the proposed framework. Following \cite{deng2024spotformer}, We first divide the input video $V$ into $T$ overlapping snippets. Each snippet is a short sequence of consecutive frames that provides the temporal context of a single frame, which is essential for distinguishing MaEs from MEs based on their duration differences. We set the snippet length to 17 and pad the beginning and end of the video with 8 repetitions of the first and last frames, respectively. This ensures that the original first and last frames can be the center frames of the first and last snippets. Within each snippet, we calculate the MDMO optical flow~\cite{liu2015main} between the first frame and each subsequent frame to obtain optical flow features.
SpotFormer~\cite{deng2024spotformer} is then employed as the feature extractor, which embeds the optical flow features into feature vectors. These vectors are concatenated along the channel dimension, resulting in $\mathbf{F}\in \mathbb{R}^{T\times D}$, where $D$ denotes the dimensionality of each snippet feature. The extracted optical flow features and the model architecture are designed explicitly for FES task.

After obtaining the optical flow features, we input the embedded feature vectors into our proposed two-branch framework to estimate the class-agnostic expression intensity scores $\bm{a}\in \mathbb{R}^{T}$ and the class-aware apex scores $\mathbf{S}\in \mathbb{R}^{T\times C}$, where $C$ represents the number of expression classes (i.e., MaE and ME). Unlike general two-branch frameworks, we decouple the two branches and process them independently during both training and inference, rather than fusing their outputs.
Since the FES task does not involve emotion recognition, the estimated intensity scores and action scores are independent of specific emotional categories. 

The overall framework further incorporates the proposed GIM module for soft pseudo-label generation and the IAC loss for contrastive learning. Note that the SpotFormer~\cite{deng2024spotformer} is only used as the feature extraction backbone due to its strong optical-flow-based representation learning capability for facial expression analysis. The proposed problem setting and overall framework design are not inherently tied to a specific backbone architecture. The GIM module operates on frame-level feature embeddings and assigns soft pseudo-labels based on feature distances, without imposing any constraints on how these features are extracted. Therefore, the proposed framework can be readily integrated with alternative backbone networks, including appearance-based or multimodal models, as well as other spatiotemporal models, that are capable of producing snippet-based frame-level features.

\subsection{Gaussian-based instance-adaptive intensity modeling (GIM)}
\label{gimsection}
In this section, we introduce a novel soft pseudo-labeling method to train the class-agnostic expression intensity branch, which models the instance-level expression intensity distribution. The proposed solution is based on the assumption that the expression intensity within each expression instance follows a smooth Gaussian distribution, with the apex frame corresponding to the peak intensity and decreasing on both sides. Neutral frames are assumed to have an intensity of 0.

To capture the intensity distribution of each expression instance, we construct instance-adaptive Gaussian distributions using the intermediate feature representations $\mathbf{X}\in \mathbb{R}^{T\times D}$ from the class-agnostic expression intensity branch and the corresponding output intensity scores $\bm{a}$.

The algorithm for constructing the instance-adaptive Gaussian distributions and assigning soft pseudo-labels is described as follows.\par
\textbf{Step 1.} Given the output intensity scores $\bm{a}$ and a point label $p_i$ with expression class $c$, we identify the pseudo-apex frame $\mathbf{v}_i^{\mathrm{apex}}$ as the frame with the highest intensity score within a predefined range $I_i$:
\begin{equation}
    \mathbf{v}_i^\mathrm{apex} = \argmax_{j \in I_i} a_j,
\label{eq0}
\end{equation}
where $I_i = \{n \in \mathbb{Z} \mid p_i-\frac{k_c}{4} \leq n \leq p_i+\frac{k_c}{4} \}$, and $k_c$ denotes the general duration of the $c$-th class expression instance (i.e., MaE or ME). The intermediate feature of the pseudo-apex frame, $\bm{x}_i^{\mathrm{apex}}$, is selected as the $\bm{\mu}_i$ for the $i$-th Gaussian distribution $g_i$:
\begin{equation}
    \bm{\mu}_i = \bm{x}_i^{\mathrm{apex}}.
\label{eq1}
\end{equation}
This selection strategy guarantees that the pseudo-apex frame acts as the center of the Gaussian distribution and possesses the highest soft pseudo-label.

\textbf{Step 2.} Centered on the pseudo-apex frame $\mathbf{v}_i^{\mathrm{apex}}$, we estimate the rough duration $L_i$ of the $i$-th expression instance by selecting neighboring expression frames whose intensity scores exceed a threshold $\theta$:
\begin{equation}
    L_i = \left|\left\{j \in J_i \mid a_j > \theta \right\}\right|,
\label{eq2}
\end{equation}
where $J_i = \{n \in \mathbb{Z} \mid \mathbf{v}_i^{\mathrm{apex}}-\frac{k_c}{2} \leq n \leq \mathbf{v}_i^{\mathrm{apex}}+\frac{k_c}{2} \}$, and $a_j$ represents the intensity score of frame $v_j$. We then expand the rough duration by a coefficient $\delta$ to account for unreliable low-intensity expression frames, thereby completing the expression proposal. For each expression proposal, we measure the feature distance between each frame and the pseudo-apex frame, and subsequently calculate the variance $\sigma_i$ for the Gaussian distribution. The formulation is as follows:
\begin{equation}
    \sigma_i = \sqrt{\frac{1}{\delta L_i}\sum_{j \in K_i}\|\bm{x}_j-\boldsymbol{\mu}_i\|_2^2},
\label{eq3}
\end{equation}
where $\|\cdot\|_2$ denotes the Euclidean distance, and $K_i = \{n \in \mathbb{Z} \mid \mathbf{v}_i^\mathrm{apex}-\frac{\delta L_i}{2} \leq n \leq \mathbf{v}_i^\mathrm{apex}+\frac{\delta L_i}{2} \}$.

\textbf{Step 3.} Finally, we build an unnormalized Gaussian distribution $g_i$ for the $i$-th expression instance:
\begin{equation}
    g_i(\bm{x}_j; \boldsymbol{\mu}_i, \sigma_i) = \exp\left(-\frac{\|\bm{x}_j-\boldsymbol{\mu}_i\|_2^2}{2\sigma_i^2}\right), \quad j \in K_i.
\label{eq4}
\end{equation}
Given the Gaussian distribution $g_i$, we can assign a soft pseudo-label to each pseudo-expression frame in the range of $(0, 1]$.

Although the Gaussian distribution is constructed as a symmetric intensity pattern centered at the apex frame, it serves as a structural prior rather than a rigid temporal assumption. The soft pseudo-labels are determined by the feature distance between each frame and the pseudo-apex frame instead of the temporal distance, which allows them to naturally form asymmetric intensity trajectories adaptive to each individual expression instance (analysis and visualizations are provided in Section~\ref{sec:pseudolabeling} with Table~\ref{strategies} and Section~\ref{sec:qualitative} with Fig.~\ref{label}). 
This effectively decouples expression intensity estimation from strict temporal symmetry assumptions, thereby allowing the GIM module to remain robust to asymmetric expression dynamics while preserving the interpretability of the Gaussian-based prior, that is, the apex frame corresponds to the peak intensity, and the intensity smoothly decreases on each side.

Note that the class-agnostic expression intensity branch is termed \textit{class-agnostic} because the output expression intensity is independent of class labels. Class information is only used during training for soft pseudo-labeling (via $k_c$), whereas the inference output is shared by both MaEs and MEs. 

\subsection{Intensity-aware contrastive learning on pseudo-labeled frames}
To further suppress neutral noise, highlight expression frames, and learn inter-class differences for action classification, we introduce contrastive learning \cite{khosla2020supervised} to pseudo-labeled frames. Additionally, we consider the impact of intensity differences and propose an Intensity-Aware Contrastive (IAC) loss. The intuition is that the intensity differences between frames are independent of the class (i.e., MaE, ME, and neutral), and we should consider these intensity differences when building contrasts on pseudo-labeled frames. For example, if two ME frames exhibit a large intensity difference (e.g., one is an onset frame and the other is an apex frame), it is not necessary to pull them closer in the feature space, even though they belong to the same class. Conversely, if an ME onset frame and a neutral frame have only a small intensity difference, it may not be appropriate to push them far apart in the feature space, even though they belong to different classes. These considerations motivate the design of the proposed IAC loss, which explicitly incorporates intensity information into feature learning rather than relying solely on class labels.

In the previous section, we focused on mining pseudo-expression frames and assigning soft pseudo-labels to them. To introduce contrastive learning, we now need to mine neutral frames.
Suppose we assign pseudo-expression labels to $N_{\mathrm{exp}}$ frames. We employ the top-$k$ strategy to select pseudo-neutral frames from those not given pseudo-labels with the top-$k$ lowest expression intensity scores. The number of pseudo-neutral frames $N_{\mathrm{neut}}$ is determined by:
\begin{equation}
    N_{\mathrm{neut}} = \min(N_{\mathrm{exp}}, T-N_{\mathrm{exp}}).
\label{eq5}
\end{equation}

Next, we select reliable pseudo-expression frames with pseudo-intensity labels greater than 0.5 and construct intensity-aware contrasts between reliable pseudo-neutral and pseudo-expression frames. Let $\mathcal{I}$ represent the set of reliable pseudo-expression and pseudo-neutral frames. The loss function is then formulated as follows:
\begin{equation}
\begin{aligned}
\mathcal{L}_{\mathrm{IAC}}\!=\!\sum_{i\in \mathcal{I}}\frac{-1}{|Q(i)|}\!\sum_{q\in Q(i)}\!\log\frac{w_{i,q}\exp(\bm{f}_i^\top \bm{f}_q/\tau)}{\sum_{e\in E(i)}w_{i,e}\exp(\bm{f}_i^\top \bm{f}_e/\tau)},
\label{eq6}
\end{aligned}
\end{equation}

\begin{equation}
\begin{aligned}
    w_{i,j}=\left\{ \begin{array}{ll}
  1-|\hat{a}_i-\hat{a}_j|, & \text{if } \widetilde{y}_i=\widetilde{y}_j \\
  |\hat{a}_i-\hat{a}_j|, & \text{if } \widetilde{y}_i\neq\widetilde{y}_j
\end{array} \right.,
\label{eq7}
\end{aligned}
\end{equation}
where $E(i) \coloneqq \mathcal{I}\textbackslash i$, and $Q(i) \coloneqq \{q\in E(i) \mid \widetilde{y}_q=\widetilde{y}_i\}$ represents the set of samples in the video who has the same pseudo-class label with the $i$-th sample, $\bm{f}_i$ is the embedded feature of the $i$-th sample (the $i$-th snippet feature of $\mathbf{F}\in \mathbb{R}^{T\times D}$), $\tau \in \mathbb{R}^+$ is a scalar temperature parameter, $\hat{a}_i$ represents the pseudo-intensity label of the $i$-th sample, respectively.

\subsection{Training and inference}
\subsubsection{Loss function}
\label{sec:lossfunction}
General P-TAL methods \cite{lee2021learning, zhang2024hr} model the class-agnostic branch as a binary classification task and optimize it using the binary cross-entropy (BCE) loss function. In contrast, this paper assumes that the expression intensity score for each expression instance follows a smooth Gaussian distribution rather than a Bernoulli distribution. Therefore, we treat it as a regression task and apply the mean squared error (MSE) loss to optimize the expression intensity branch:
\begin{equation}
\begin{aligned}
    \mathcal{L}_{\mathrm{GIM}}=\frac{1}{N_{\mathrm{neut}}+N_{\mathrm{exp}}}\sum_{i\in \mathcal{T}}(a_i - \hat{a}_i)^2,
\label{eq8}
\end{aligned}
\end{equation}
where $\mathcal{T}$ represents the set of all pseudo-labeled frames, and $a_i$ and $\hat{a}_i$ represent the output intensity score and the corresponding soft pseudo-label of the frame $\mathbf{v}_i$, respectively.

Following the previous work \cite{hong2021cross}, due to the sparsity of facial expressions in the video, we add an L1 normalization loss on the intensity scores:
\begin{equation}
\begin{aligned}
    \mathcal{L}_{\mathrm{norm}}=\|\bm{a}\|_1,
\label{eq9}
\end{aligned}
\end{equation}
where $\|\cdot\|_1$ is a L1-norm function.

Due to $\mathcal{L}_{\mathrm{norm}}$ and the tendency for most frames to have lower scores (neutral and low-intensity expression frames), the model tends to produce low expression intensity scores. Therefore, we encourage reliable pseudo-expression frames to generate higher intensity scores:
\begin{equation}
\begin{aligned}
    \mathcal{L}_{\mathrm{reward}}=\frac{1}{|S|}\sum_{i\in S}-|a_i|,
\label{eq9}
\end{aligned}
\end{equation}
where $S$ represents the set of reliable pseudo-expression frames whose soft intensity label is greater than 0.5.

We also employ a video smooth loss to encourage temporal consistency in video output by ensuring that consecutive frames have similar predictions, stabling the training process:
\begin{equation}
\begin{aligned}
    \mathcal{L}_{\mathrm{smooth}}=\frac{1}{T-1} \sum_{t=1}^{T-1} \| a_{t+1} - a_t \|_1.
\label{eq10}
\end{aligned}
\end{equation}

Unlike general P-TAL frameworks that fuse the outputs of two branches for action classification, we process each branch independently. 
However, after re-designing the overall framework to remove branch fusion, we found that the original positive sample mining strategy, which selected all reliable expression frames as positive samples for training the action classification branch, led to an excessive number of positive samples. This, in turn, inflated the overall action scores and generated redundant false positive proposals. Therefore, in this paper, we revise the action classification branch to a class-aware apex classification branch, which focuses on distinguishing MaEs and MEs based on their pseudo-apex frames. Specifically, we mark only a limited set of pseudo-apex frames as positive samples. This is because the optical flow features extracted at the apex frames provide the most critical information about a facial expression instance in terms of both intensity and duration~\cite{deng2024spotformer}.
Specifically, for positive samples, we select only the pseudo-apex frames whose soft pseudo-labels are 1, along with their neighboring $m_c$ frames. For negative samples, we include all pseudo-neutral frames and pseudo-expression frames (excluding the positive samples) whose soft pseudo-labels are greater than 0.5, together with their surrounding $m_{\mathrm{neut}}$ frames. We employ Focal loss \cite{lin2017focal} since there is a significant data imbalance between positive and negative samples, which can be formulated as:

\begin{flalign}
    \mathcal{L}_{\mathrm{apex}} =& -\frac{1}{|\mathcal{I}_c^+|}\sum_{i \in \mathcal{I}_c^+}\sum_{c=1}^{C}\alpha(1-s_{i,c})^\gamma \log s_{i,c} \\ \nonumber
    &- \frac{1}{|\mathcal{I}_c^-|}\sum_{i \in \mathcal{I}_c^-} \sum_{c=1}^{C}(1-\alpha)s_{i,c}\log (1 - s_{i,c}),
\label{eq12}
\end{flalign}
where $\mathcal{I}_c^+$ and $\mathcal{I}_c^-$ represent the set of positive and negative frames of the $c$-th class, $\alpha$ and $\gamma$ are hyperparameters, respectively.

Finally, the total loss function can be summarized as:
\begin{flalign}
    \mathcal{L}=&\mathcal{L}_{\mathrm{GIM}} + \mathcal{L}_{\mathrm{apex}} + \mathcal{L}_{\mathrm{reward}} \\ \nonumber
    &+ \lambda_1\mathcal{L}_{\mathrm{smooth}} + \lambda_2\mathcal{L}_{\mathrm{norm}} + 
    \lambda_3\mathcal{L}_{\mathrm{IAC}},
\label{eq13}
\end{flalign}
where $\lambda_1$, $\lambda_2$, and $\lambda_3$ are hyper-parameters for balancing the losses, which are determined empirically.

\subsubsection{Training pipeline}
In the early stages of training, the output expression intensity scores may not accurately reflect the true expression intensity, and the highest score does not surely indicate the apex frame. To address this, we adopt an easy-to-hard learning strategy and incorporate several warm-up epochs to ensure that the output intensity score reliably represents the expression intensity. Specifically, 1) In the first stage, we assign hard pseudo-labels to neighboring frames around each labeled frame $p_i$ within the range $[p_i - k_{s1}, p_i + k_{s1}]$. 2) In the second stage, we construct a Gaussian distribution centered on the labeled frame and assign soft pseudo-labels within a small predefined range $[p_i - k_{s2}, p_i + k_{s2}]$. 3) In the third stage, we apply our proposed GIM for soft pseudo-labeling and model training. The pseudo-apex frame and the soft pseudo-labeling range are updated at each epoch to enhance intensity-related feature learning.

\subsubsection{Inference}
\label{sec:inference}
During the inference phase, we first obtain the expression intensity scores $\bm{a}$ for candidate proposal generation and the class-aware apex scores $\mathbf{S}$ for determining the expression class. Note that, unlike general P-TAL frameworks that fuse the outputs of two branches for class determination, our framework relies solely on class-aware apex scores. Then, we apply a multi-threshold strategy on the expression intensity scores $\bm{a}$ to generate candidate expression proposals, where each proposal includes consecutive frames with intensity scores higher than a given threshold. Each proposal is represented as ($s_i$, $e_i$, $c_i$, $p_i^{\mathrm{OIC}}$), where $s_i$, $e_i$, $c_i$, and $p_i^{\mathrm{OIC}}$ represent the onset frame, offset frame, expression type, and the outer-inner-contrastive (OIC) score \cite{shou2018autoloc}, respectively.
Specifically, $c_i$ is determined by applying a threshold of 0.5 to the apex score of the pseudo-apex frame since it is the most representative frame for each expression instance. The OIC score can be calculated as follows:
\begin{equation}
\begin{aligned}
    p_i^{\mathrm{OIC}}= \frac{1}{L_i} \sum_{t = s_i}^{e_i} a_t - \frac{1}{\frac{L_i}{2}}\left(\sum_{t=s_i- \frac{L_i}{4}}^{s_i - 1} a_t + \sum_{t=e_i + 1}^{e_i + \frac{L_i}{4}} a_t \right),
\label{eq15}
\end{aligned}
\end{equation}
\begin{equation}
\begin{aligned}
    L_i = e_i - s_i + 1.
\label{eq16}
\end{aligned}
\end{equation}

Finally, we apply class-wise NMS \cite{bodla2017soft, yin2023aware} to remove redundant ones. Note that each proposal is independently evaluated for both MaE and ME classification, and thus the inference stage may naturally produce partially overlapping MaE and ME proposals. Specifically, if a proposal falls within the general duration of a MaE and the MaE apex score of its pseudo-apex frame exceeds the pre-defined threshold 0.5, the proposal is classified as an MaE. Conversely, if a proposal falls within the general duration of an ME and the ME apex score of its pseudo-apex frame exceeds the threshold 0.5, the proposal is classified as an ME.

\begin{table*}[tbp]
\setlength\tabcolsep{4pt}
  \centering
  \caption{Comparison with the state-of-the-art methods on SAMM-LV and CAS(ME)$^2$ in terms of F1 score. The dagger \dag denotes the method originally for P-TAL and reproduced for P-FES. “Modality” indicates the input type for each method: “k-step OF” refers to optical flow extracted with a temporal step of k between frames; “RGB” refers to raw image inputs; “k-step RGB” refers to raw images sampled every k frames; “MDMO” denotes MDMO~\cite{liu2015main} features between adjacent frames; and “SW-MRO” denotes sliding window–based multi-resolution optical flow features proposed in \cite{deng2024spotformer}.}
    \begin{tabular*}{0.7\hsize}{@{}@{\extracolsep{\fill}}l|l|c|ccc|ccc@{}}
    \toprule
    \multicolumn{2}{c|}{Methods} &Modality& \multicolumn{3}{c|}{SAMM-LV} & \multicolumn{3}{c}{CAS(ME)$^2$} \\
    \multicolumn{2}{c|}{}&&MaE&ME&Overall&MaE&ME&Overall \\
    \midrule
    F-FES&
    SOFTNet~\cite{liong2021shallow}&k-step OF&0.2169&0.1520&0.1881&0.2410&0.1173&0.2022\\
    &Concat-CNN~\cite{yang2021facial}&RGB&0.3553&0.1155&0.2736&0.2505&0.0153&0.2019\\
    &LSSNet~\cite{yu2021lssnet}&RGB+OF&0.2810&0.1310&0.2380&0.3770&0.0420&0.3250\\
    &3D-CNN~\cite{yap20223d}&k-step RGB&0.1595&0.0466&0.1084&0.2145&0.0714&0.1675\\
    &MTSN~\cite{liong2022mtsn}&k-step OF&0.3459&0.0878&0.2867&0.4104&0.0808&0.3620\\
    &ABPN~\cite{leng2022abpn}&MDMO&0.3349&0.1689&0.2908&0.3357&0.1590&0.3117\\
    &AUW-GCN~\cite{yin2023aware}&MDMO&0.4293&0.1984&0.3728&0.4235&0.1538&0.3834\\
    &SpoT-GCN~\cite{deng2024multi}&SW-MRO&0.4631&0.4035&0.4454&0.4340&0.2637&0.4154\\
    &SpotFormer~\cite{deng2024spotformer}&SW-MRO&0.4447&0.4281&0.4401&0.5061&0.2817&0.4841\\ 

    \midrule
    P-FES
    &LAC\cite{lee2021learning}\dag&SW-MRO&0.3714&0.1983&0.3223&0.3889&0.0833&0.3598\\
    &HR-Pro\cite{zhang2024hr}\dag&SW-MRO&0.3395&0.1667&0.2895&0.3515&0.1345&0.3261\\
    &TSP-Net\cite{xia2024realigning}\dag&SW-MRO&0.3152&0.1567&0.2703&0.3781&0.0571&0.3358\\
    &\textbf{Deng \textit{et al.}~\cite{deng2025gaussianbased}}&SW-MRO&\textbf{0.4189}&0.2033&0.3587&\textbf{0.4395}&0.0588&0.4000\\
&\textbf{Ours}&SW-MRO&0.4176&\textbf{0.2417}&\textbf{0.3705}&0.4339&\textbf{0.1370}&\textbf{0.4023}\\
    \bottomrule
    \end{tabular*}%
  \label{results1}%
\end{table*}%
\begin{table}[tbp]
\setlength\tabcolsep{7pt}
  \centering
  \caption{Comparison with the state-of-the-art methods on CAS(ME)$^3$ in terms of F1 score. The dagger \dag denotes the method originally for P-TAL and reproduced for P-FES.}
    \begin{tabular}{l|l|ccc}
    \toprule
    \multicolumn{2}{c|}{Methods} & \multicolumn{3}{c}{CAS(ME)$^3$}\\
    \multicolumn{2}{c|}{}&MaE&ME&Overall\\
    \midrule
    F-FES&
    LGSNet~\cite{yu2023lgsnet}&-&0.0990&0.2350\\
    &SpotFormer~\cite{deng2024spotformer}&0.2664&0.2037&0.2559\\
    \midrule
    P-FES&LAC~\cite{lee2021learning}\dag&0.2235&0.0677&0.2146\\
    &TSP-Net~\cite{xia2024realigning}\dag&0.1751&0.0029&0.1484\\
    &Deng \textit{et al.}~\cite{deng2025gaussianbased}&0.2396&0.0708&0.2273\\
&\textbf{Ours}&\textbf{0.2438}&\textbf{0.0958}&\textbf{0.2335}\\
    \bottomrule
    \end{tabular}%
  \label{resultscasme3}%
\end{table}%



\section{Experiments}
\subsection{Experimental settings}
\subsubsection{Datasets}
We first follow the protocol of MEGC2021 and mainly validate our method on two datasets: \cite{yap2020samm} and CAS(ME)$^2$ \cite{qu2017cas}. The SAMM-LV dataset has 147 annotated videos containing 343 MaE clips and 159 ME clips. The CAS(ME)$^2$ dataset has 98 annotated videos containing 300 MaE clips and 57 ME clips. The frame rate is 200 fps for the SAMM-LV dataset and 30 fps for the CAS(ME)$^2$ dataset. To align the frame rates of both datasets, we subsample every 7th frame from the SAMM-LV dataset to achieve a frame rate close to 30fps.

\noindent
\subsubsection{Evaluation metrics}
We employ the leave-one-subject-out cross-validation strategy in the experiments. An expression proposal is considered true positive (TP) if the Intersection over Union (IoU) between the expression proposal and a ground-truth expression instance satisfies:
\begin{equation}
\begin{aligned}
    \frac{W_{\mathrm{Proposal}}\cap W_{\mathrm{GroundTruth}}}{W_{\mathrm{Proposal}}\cup W_{\mathrm{GroundTruth}}}\geq\theta_{\mathrm{IoU}},
\end{aligned}
\end{equation}
where $\theta_{\mathrm{IoU}}$ is the IoU threshold, set to 0.5. We calculate the F1 score to evaluate the performance of our model and compare it with previous methods.

\noindent
\subsubsection{Training details}
\label{trainingdetails}
We generate single-frame annotations using a Gaussian distribution centered on the ground-truth apex frame for each instance. The model is trained by the Adam optimizer \cite{kingma2014adam} on both datasets for 100 epochs with a learning rate of $2.0 \times 10^{-5}$ and a weight decay of 0.1. The coefficient $\delta$ for duration estimation is set to 1.2. For the multi-stage training, the epochs for each stage are 1, 4, and 95, respectively. $k_c$ is set to 16 for ME and 32 for MaE (corresponding to a frame rate of 30 fps), respectively. $k_{s1}$ is set to 3 for MEs and 5 for MaEs, $k_{s2}$ is set to 2 for MEs and 4 for MaEs. $m_c$ is set to 1 for MEs and 2 for MaEs, and $m_{\mathrm{neut}}$ is set to 6. We set the loss weight $\lambda_*$ to 0.1, 0.3, and $2.0 \times 10^{-5}$ for SAMM-LV, and to 0.1, 2.5, and $1.4 \times 10^{-4}$ for CAS(ME)$^2$, respectively. The threshold $\theta$ for estimating the rough duration of each expression proposal decreases linearly from 0.8 to 0.5 over 30 epochs and then remains at 0.5 until the end.

\begin{table*}[tbp]
\setlength\tabcolsep{4pt}
  \centering
  \caption{Validation of apex frame detection on SAMM-LV, CAS(ME)$^2$, and CAS(ME)$^3$ in terms of NMSE. The dagger \dag denotes the method originally for P-TAL and reproduced for P-FES.}
    \begin{tabular*}{0.8\hsize}{@{}@{\extracolsep{\fill}}l|l|ccc|ccc|ccc@{}}
    \toprule
    \multicolumn{2}{c|}{Methods} & \multicolumn{3}{c|}{SAMM-LV} & \multicolumn{3}{c|}{CAS(ME)$^2$}&\multicolumn{3}{c}{CAS(ME)$^3$} \\
    \multicolumn{2}{c|}{}&MaE&ME&Overall&MaE&ME&Overall&MaE&ME&Overall\\
    \midrule
    F-FES&SpotFormer~\cite{deng2024spotformer}&0.2738&0.0961&0.2254&0.1345&0.0836&0.1315&0.1111&0.0766&0.1077\\

    \midrule
    P-FES&LAC\cite{lee2021learning}\dag&0.2307&0.2473&0.2348&0.1995&0.0000&0.1967&0.2290&0.4931&0.2312\\
    &Deng \textit{et al.}~\cite{deng2025gaussianbased}&0.2145&\textbf{0.2011}&0.2127&0.2015&0.2587&0.2023&0.2286&0.2780&0.2301\\
&\textbf{Ours}&\textbf{0.2097}&0.2241&\textbf{0.2118}&\textbf{0.1916}&\textbf{0.2062}&\textbf{0.1921}&\textbf{0.2170}&\textbf{0.2607}&\textbf{0.2180}\\
    \bottomrule
    \end{tabular*}%
  \label{resultsapex}%
\end{table*}%

\begin{table*}[h]
\setlength\tabcolsep{4pt}
  \centering
  \caption{Ablation study on the two-branch framework optimization strategy. `Fusion' denotes the general two-branch frameworks that fuse the outputs of both branches during training and inference.
`Decoupled-Action' denotes a decoupled framework that processes the two branches independently while keeping the general action classification branch for MaE and ME classification.
`Decoupled-Apex' denotes our proposed framework, which also decouples the two branches but replaces the general action classification branch with the class-aware apex classification branch for MaE and ME classification.}
    \begin{tabular*}{0.75\hsize}{@{}@{\extracolsep{\fill}}l|ccc|ccc|ccc@{}}
    \toprule
    &\multicolumn{3}{c|}{SAMM-LV}&\multicolumn{3}{c|}{CAS(ME)$^2$}&\multicolumn{3}{c}{CAS(ME)$^3$}\\
    Strategy&MaE&ME&Overall&MaE&ME&Overall&MaE&ME&Overall\\
    \midrule
    Fusion~\cite{deng2025gaussianbased}&\textbf{0.4189}&0.2033&0.3587&\textbf{0.4395}&0.0588&0.4000&0.2396&0.0708&0.2273\\
    Decoupled-Action&0.3961&0.1613&0.3358&0.3936&0.1277&0.3618&0.1960&0.0638&0.1812\\
    \textbf{Decoupled-Apex (Ours)}&0.4176&\textbf{0.2417}&\textbf{0.3705}&0.4339&\textbf{0.1370}&\textbf{0.4023}&\textbf{0.2438}&\textbf{0.0958}&\textbf{0.2335}\\
    \bottomrule
    \end{tabular*}%
  \label{frameworkablation}%
\end{table*}%

\subsection{Comparison with state-of-the-art methods}
We first compare the performance with state-of-the-art (SOTA) deep learning methods, and the results are shown in Table~\ref{results1}. We reproduce several SOTA P-TAL methods to the P-FES task for the comparison. For a fair comparison, we use the same optical flow extraction, feature extraction method, and post-processing process as in our method when reproducing these SOTA P-TAL methods, which significantly improves the performance of P-FES. In addition, we fine-tune the hyperparameters when reproducing these methods, such as the learning rate, to achieve optimal performance. It can be seen that our method outperforms SOTA point-supervised methods by 14.96\% on SAMM-LV and 11.81\% on CAS(ME)$^2$. We also present the results of several F-FES methods for comparison, which demonstrate that our method achieves competitive performance in MaE spotting but relatively low performance in ME spotting, all with a very limited annotation cost. The reason is that our method focuses on significantly suppressing neutral noise, which may overshadow extremely subtle MEs without the help of precise frame-level annotations. 

We also compare our method with the previous conference version, and the results show that it outperforms the earlier approach by 18.89\% and 132.99\% in ME spotting on SAMM-LV and CAS(ME)$^2$, respectively. This improvement arises from addressing the limitations of directly following the general P-TAL framework, where the performance of ME spotting tends to be lower due to several factors. First, MEs generally exhibit lower intensity. Second, combining action scores and intensity scores for action classification requires optimizing two branches simultaneously, making it difficult for both outputs to reach their optimum. Third, during post-processing, action scores are suppressed when fused with intensity scores, and the relatively low intensity scores of MEs further reduce their final scores. As a result, ME spotting performance is negatively affected. The redesigned framework optimizes the two branches separately with individual loss functions, ensuring that action scores remain independent of intensity scores. This strategy also allows us to mine more negative samples for class-aware apex classification.

In addition, we further conduct experiments on a large scale dataset CAS(ME)$^3$ \cite{li2022cas} to validate the effectiveness of our model. Following Yu \textit{et al.} \cite{yu2023lgsnet} and Deng \textit{et al.} \cite{deng2024spotformer}, we pre-process the dataset by only considering ground-truth MEs whose length are less than 15 frames and ground-truth MaEs with a length between 16 and 120 frames for training and validation to ensure a fair comparison. After pre-processing, we obtain 2231 ground-truth MaEs and 285 ground-truth MEs for CAS(ME)$^3$. It is worth noting that the performance of our method on the large-scale dataset CAS(ME)$^3$, shown in Table \ref{resultscasme3}, is close to SOTA F-FES methods, which validates the generalization ability and robustness of our model.

\subsection{Apex frame detection}
One of the advantages of our proposed GIM is that it can naturally detect the apex frame within each facial expression instance. Although the FES task itself does not explicitly require apex frame detection, accurately identifying the apex frame is essential for understanding the intensity changes of facial expressions and for improving recognition accuracy in subsequent downstream tasks. In ME recognition, most advanced methods~\cite{li2025fed, wang2025pme, li2024structure, zhai2023feature} rely on optical flow features or frame differences between the onset and apex frames as motion cues to capture subtle facial movements. In MaE recognition, many methods~\cite{zhao2025enhancing, zhang2023weakly, lan2025expllm, jin2024facial} rely on static images that effectively represent the expressive moment for classification. Therefore, we propose to evaluate the performance of apex frame detection in the FES task, which not only verifies the intensity modeling capability of our method but also demonstrates its potential for downstream tasks such as ME and MaE recognition.

Following \cite{li2020joint}, we calculate the normalized mean absolute error (NMAE) to report the effectiveness of the apex frame detection method, which can be described as:
\begin{equation}
\begin{aligned}
    NMAE=\frac{1}{K}\sum_{i=1}^K\hat{e}_i,
\end{aligned}
\end{equation}
\begin{equation}
\begin{aligned}
    \hat{e}_i=\frac{e_i}{l_i},
\end{aligned}
\end{equation}
where $e_i$ represents the frame distance between the detected apex frame and the ground-truth apex frame of the $i$-th sample. $l_i$ is the ground-truth length of the $i$-th expression instance, and $K$ represents the number of TP samples. Note that we only consider TP when evaluating the performance of apex frame detection.
As shown in Table \ref{resultsapex}, our method achieves 9.80\% and 2.34\% reductions compared to the previous method on the SAMM-LV and CAS(ME)$^2$ datasets, respectively. The results demonstrate that our method can detect the apex frame more accurately, which is highly significant for subsequent tasks.

\begin{table*}[t]
\setlength\tabcolsep{4pt}
  \centering
  \caption{Ablation study on pseudo-labeling strategies.}
    \begin{tabular*}{0.65\hsize}{@{}@{\extracolsep{\fill}}l|ccc|ccc|ccc@{}}
    \toprule
    &\multicolumn{3}{c|}{SAMM-LV}&\multicolumn{3}{c|}{CAS(ME)$^2$}&\multicolumn{3}{c}{CAS(ME)$^3$}\\
    Strategy&MaE&ME&Overall&MaE&ME&Overall&MaE&ME&Overall\\
    \midrule
    Hard&0.3882&0.0923&0.3194&0.4160&\textbf{0.1493}&0.3902&0.2134&0.0677&0.1972\\
    Soft&0.2888&0.2303&0.2671&0.3927&0.0909&0.3515&0.2266&0.0352&0.2139\\
    Class-wise&0.2050&0.0894&0.1597&0.2857&0.0602&0.2406&0.1145&0.0229&0.1040\\
    Ours&\textbf{0.4176}&\textbf{0.2417}&\textbf{0.3705}&\textbf{0.4339}&0.1370&\textbf{0.4023}&\textbf{0.2438}&\textbf{0.0958}&\textbf{0.2335}\\
    \bottomrule
    \end{tabular*}%
  \label{strategies}%
\end{table*}%

\begin{table*}[h]
\setlength\tabcolsep{4pt}
  \centering
  \caption{Ablation study on pseudo-labeling strategies for overlapping Gaussian distributions.}
    \begin{tabular*}{0.75\hsize}{@{}@{\extracolsep{\fill}}l|ccc|ccc|ccc@{}}
    \toprule
    &\multicolumn{3}{c|}{SAMM-LV}&\multicolumn{3}{c|}{CAS(ME)$^2$}&\multicolumn{3}{c}{CAS(ME)$^3$}\\
    Strategy&MaE&ME&Overall&MaE&ME&Overall&MaE&ME&Overall\\
    \midrule
    Higher&0.3988&0.2213&0.3523&0.4187&\textbf{0.1772}&0.3914&0.2292&0.0520&0.2203\\
    Lower&0.4038&0.1833&0.3436&0.4126&0.1194&0.3818&0.2274&0.0400&0.2165\\
    \textbf{Random (Ours)}&\textbf{0.4176}&\textbf{0.2417}&\textbf{0.3705}&\textbf{0.4339}&0.1370&\textbf{0.4023}&\textbf{0.2438}&\textbf{0.0958}&\textbf{0.2335}\\
    \bottomrule
    \end{tabular*}%
  \label{overlapGau}%
\end{table*}%

\subsection{Ablation study}


\subsubsection{Two-branch framework optimization strategy}
In this subsection, we make a direct comparison with the general two-branch framework to verify the effectiveness of our proposed framework. The quantitative results are summarized in Table~\ref{frameworkablation}.
As discussed earlier, `Fusion' denotes the general two-branch framework that fuses the outputs of both branches during training and inference. This fusion causes a decrease in ME spotting performance, as MEs usually exhibit lower intensity than MaEs, leading to their action scores being suppressed after fusion.
After removing the fusion mechanism (`Decoupled–Action'), if we still adopt the original action classification branch that treats all reliable expression frames as positive samples, the number of positive samples becomes excessive. This leads to inflated overall action scores, an increased number of false positive proposals, and consequently degraded spotting performance.
Our `Decoupled–Apex' framework, which replaces the original action classification branch with the proposed class-aware apex classification branch, is more suitable for the separate two-branch setting. By focusing on the pseudo-apex frames for classification, it effectively mines more discriminative negative samples and improves the balance between positive and negative supervision, achieving the best overall performance.

\subsubsection{Pseudo-labeling strategy}
\label{sec:pseudolabeling}
To verify the effectiveness of our proposed GIM, we conduct ablation studies on various pseudo-labeling strategies. In Table~\ref{strategies}, \textbf{Hard} denotes our implemented hard pseudo-labeling method, in which the proposed GIM is used to estimate the duration of each expression instance and to assign hard pseudo-labels accordingly, \textbf{Soft} refers to using cosine similarity in the feature space as the soft pseudo-label instead of employing the proposed GIM, and \textbf{Class-wise} indicates that we use the class-wise average feature of point labels as the $\mu$ of the Gaussian distribution instead of detecting the pseudo-apex frame. The comparison with \textbf{Hard} demonstrates the effectiveness of using a soft pseudo-labeling strategy for modeling the intensity of expression frames, while the comparison with \textbf{Soft} highlights both the superiority and robustness of our proposed GIM in assigning soft pseudo-labels. In addition, the comparison with \textbf{Class-wise} emphasizes the effectiveness of our instance-adaptive method as well as the importance of selecting an appropriate $\mu$ when building Gaussian distributions. This is because even expressions within the same class can vary greatly in intensity. Using class-averaged features corresponding to point labels as $\mu$ for the Gaussian distribution would cause most expression frames to deviate significantly from $\mu$, resulting in lower pseudo-intensity labels and consequently harming performance. 
Moreover, in \textbf{Class-wise}, we rely on point labels as pseudo-apex frames for class-aware apex classification. This inaccuracy also affects model training.
-Pro
In summary, the results demonstrate the effectiveness of our proposed GIM module. The Gaussian distribution is constructed based on the intensity-related feature distance between each frame and the pseudo-apex frame, making it adaptive to individual expression instances. For any given frame, a larger distance from the pseudo-apex frame in feature space corresponds to a lower intensity label. This formulation ensures that the intensity assignment depends on feature distance rather than temporal distance, thereby capturing expression-specific dynamics. 

\subsubsection{Handling Overlapping Gaussian Distributions}

This subsection investigates the special case where two generated Gaussian distributions overlap. We distinguish between two scenarios: (1) consecutive instances of the same class (MaEs or MEs), and (2) consecutive instances of different classes.

\begin{table*}[tbp]
\setlength\tabcolsep{3pt}
  \centering
  \caption{Ablation study on loss functions.}
    \begin{tabular*}{0.9\hsize}{@{}@{\extracolsep{\fill}}ccccc|ccc|ccc|ccc@{}}
    \toprule
    \multicolumn{5}{c|}{Loss functions}&\multicolumn{3}{c|}{SAMM-LV}&\multicolumn{3}{c|}{CAS(ME)$^2$}&\multicolumn{3}{c}{CAS(ME)$^3$}\\
    $\mathcal{L}_{\mathrm{GIM}}$(MSE)&$\mathcal{L}_{\mathrm{GIM}}$(BCE)&$\mathcal{L}_{\mathrm{IAC}}$&$\mathcal{L}_{\mathrm{con}}$&Others&MaE&ME&Overall&MaE&ME&Overall&MaE&ME&Overall\\
    \midrule
    &&\Checkmark&&\Checkmark&0.3882&0.0923&0.3194&0.4160&0.1493&0.3902&0.2134&0.0677&0.1972\\
    &\Checkmark&\Checkmark&&\Checkmark&0.2783&0.1415&0.2378&0.3134&0.0556&0.2801&0.2241&0.0127&0.2139\\
    \Checkmark&&&&\Checkmark&0.4107&0.2016&0.3543&0.3988&\textbf{0.1791}&0.3783&0.2186&0.0593&0.2111\\
    \Checkmark&&&\Checkmark&\Checkmark&0.4040&0.2083&0.3542&0.4114&0.1791&0.3891&0.2227&0.0722&0.2132\\

    \Checkmark&&\Checkmark&&\Checkmark&\textbf{0.4176}&\textbf{0.2417}&\textbf{0.3705}&\textbf{0.4339}&0.1370&\textbf{0.4023}&\textbf{0.2438}&\textbf{0.0958}&\textbf{0.2335}\\
    \bottomrule
    \end{tabular*}%
  \label{loss}%
\end{table*}%

\begin{table*}[h]
\setlength\tabcolsep{4pt}
  \centering
  \caption{Hyper-parameter analysis on $k_c$.}
    \begin{tabular*}{0.6\hsize}{@{}@{\extracolsep{\fill}}l|l|ccc|ccc|c@{}}
    \toprule
    &&\multicolumn{3}{c|}{SAMM-LV}&\multicolumn{3}{c|}{CAS(ME)$^2$}&\\
    &$k_c$&MaE&ME&Overall&MaE&ME&Overall&Overall\\
    \midrule
    For ME&8&0.4094&0.2124&0.3529&0.4181&0.1231&0.3909&0.3691\\
    &12&0.4172&0.2131&0.3630&0.4188&\textbf{0.1449}&0.3912&0.3751\\
    &\textbf{16}&0.4176&\textbf{0.2417}&\textbf{0.3705}&\textbf{0.4339}&0.1370&\textbf{0.4023}&\textbf{0.3843}\\
    &20&\textbf{0.4339}&0.1525&0.3608&0.4197&0.1333&0.3892&0.3732\\
    \midrule
    For MaE&16&0.4129&0.2016&0.3549&0.4239&0.1449&0.3966&0.3728\\
    &\textbf{32}&0.4176&\textbf{0.2417}&\textbf{0.3705}&\textbf{0.4339}&0.1370&0.4023&\textbf{0.3843}\\
    &48&0.4149&0.2204&0.3628&0.4328&0.1429&\textbf{0.4040}&0.3807\\
    &64&\textbf{0.4224}&0.1863&0.3656&0.4238&\textbf{0.1591}&0.3910&0.3772\\
    \bottomrule
    \end{tabular*}%
  \label{kc}%
\end{table*}%

\begin{table*}[h]
\setlength\tabcolsep{4pt}
  \centering
  \caption{Hyper-parameter analysis on $m_c$. REC and PRE denote the recall and precision rates, respectively.}
    \begin{tabular*}{0.75\hsize}{@{}@{\extracolsep{\fill}}l|l|ccccc|ccccc@{}}
    \toprule
    &&\multicolumn{5}{c|}{SAMM-LV}&\multicolumn{5}{c}{CAS(ME)$^2$}\\
    &$m_c$&MaE&ME&REC&PRE&Overall&MaE&ME&REC&PRE&Overall\\
    \midrule
    For ME&0&\textbf{0.4356}&0.1400&0.3207&\textbf{0.4316}&0.3680&0.4226&0.0968&0.3754&0.4123&0.3930\\
    &\textbf{1}&0.4176&\textbf{0.2417}&0.3307&0.4213&\textbf{0.3705}&\textbf{0.4339}&0.1370&\textbf{0.3866}&\textbf{0.4195}&\textbf{0.4023}\\
    &2&0.4187&0.1972&\textbf{0.3406}&0.3677&0.3537&0.4224&\textbf{0.1600}&\textbf{0.3866}&0.4023&0.3943\\
    \midrule
    For MaE&0&0.4069&0.2302&0.2988&\textbf{0.4213}&0.3497&0.4271&0.1471&0.3669&\textbf{0.4352}&0.3982\\
    &\textbf{2}&0.4176&\textbf{0.2417}&\textbf{0.3307}&\textbf{0.4213}&\textbf{0.3705}&\textbf{0.4339}&0.1370&0.3866&0.4195&\textbf{0.4023}\\
    &4&\textbf{0.4203}&0.1583&0.3187&0.3912&0.3513&0.4156&\textbf{0.1739}&\textbf{0.3894}&0.3949&0.3921\\
    \bottomrule
    \end{tabular*}%
  \label{mc}%
\end{table*}%

\textbf{Same-class overlap.} In the first scenario, a single frame may be assigned multiple soft pseudo-labels when close point-labels are used to build partially overlapping Gaussian distributions. To address this, we assign only one label by randomly retaining one and discarding the other. This strategy prevents bias toward either high- or low-intensity labels, thereby stabilizing model training. We further compared this random selection strategy with two alternative solutions: (a) keeping the higher soft label (denoted as Higher), and (b) keeping the lower soft label (denoted as Lower). As shown in Table~\ref{overlapGau}, the random strategy achieves the best overall performance.

\textbf{Different-class overlap.} In the second scenario, consecutive instances of different classes may overlap. Although we do not conduct a separate ablation for this case, we adopt a simple solution in our implementation: two arrays are used to store pseudo-labels of different classes independently, even when they share the same intensity output. In such cases, a frame may carry two pseudo-labels for intensity supervision, and both are used in the loss computation by averaging the corresponding losses. This design ensures that the model can capture composite expressions occurring in different facial regions (e.g., a MaE in the eyebrows followed by an ME at the mouth corners).

\subsubsection{Loss function}
We conduct ablation studies on loss functions to verify the effectiveness of our proposed modules. The results are shown in Table~\ref{loss}. First, we validate the effectiveness of the proposed GIM. By comparing the results of using the MSE loss and the BCE loss, we can see that the choice of treating the expression intensity branch as a regression task performs better than treating it as a binary classification task. This is because when our GIM generates soft pseudo-labels for low-intensity expression frames, and these labels accurately describe the intensity, using the MSE loss allows them to be treated as direct intensity supervision without causing a large loss value. However, using BCE loss implies that we expect the model output to be either 0 or 1, which contradicts the fact that the intensity of different expression frames varies, making it difficult to distinguish between frames of different intensities. During training, even if the soft pseudo-labels are sufficiently accurate, the loss values can still be large, severely hindering model optimization.

The first row of Table~\ref{loss} shows the results obtained when only the proposed GIM is used for hard pseudo-labeling, which corresponds to \textbf{Hard} in Table~\ref{strategies}, thereby demonstrating the effectiveness of our proposed soft pseudo-labeling strategy for expression intensity modeling.

In addition, we evaluate the effectiveness of the proposed IAC loss. Specifically, $\mathcal{L}_{\mathrm{con}}$ represents a general contrastive loss \cite{khosla2020supervised}, designed to enhance the distinguishability of the learned features, which simply encourages positive samples to be closer in the feature space while pushing negative samples farther apart. In Table~\ref{loss}, the third, fourth, and fifth rows present quantitative results that demonstrate the effectiveness of our IAC loss, which considers the characteristics of expression intensity and ultimately leads to better feature discrimination and improved overall performance in the P-FES task.

\subsubsection{Hyper-parameters $k_c$ and $m_c$}
In this subsection, we investigate the sensitivity of two key hyper-parameters, $k_c$ and $m_c$, which are introduced in the proposed framework. Specifically, $k_c$ is the range parameter that controls the spread of the Gaussian distribution in the GIM module, while $m_c$ determines the number of neighboring frames considered during positive sample mining in the class-aware apex classification branch.

To evaluate their impact, we vary each hyper-parameter while keeping all other settings fixed. For 
$k_c$, we experiment with several values \{8, 12, 16, 20\} for ME and \{24, 32, 48, 64\} for MaE, to examine how it influences the generation of soft pseudo-labels and overall model performance. For $m_c$, we test \{0, 1, 2\} for ME and \{0, 2, 4\} for MaE to assess the influence of the temporal range of positive sampling on the classification performance.

As shown in Table~\ref{kc}, although the optimal $k_c$ varies across datasets, our method achieves the best overall performance when $k_c$ is set to 16 for ME and 32 for MaE. Therefore, we adopt this setting in all comparisons with other SOTA methods to validate the generalization ability of our method. Smaller values of $k_c$ restrict the effective range of the Gaussian distribution, leading to underestimation of expression intensity, whereas larger values tend to assign soft labels to noisy neutral frames or to frames belonging to other expression instances, thereby degrading performance.

As shown in Table~\ref{mc}, the optimal performance is achieved when $m_c$ is set to 1 for ME and 2 for MaE. When $m_c$ is smaller, i.e., fewer positive samples are selected, the model tends to produce fewer true positives, resulting in lower recall. Conversely, when $m_c$ is larger, i.e., more positive samples are considered, recall increases but precision decreases, as redundant positive samples inflate the overall scores and lead to more false positive proposals.

\begin{table*}[t]
\setlength\tabcolsep{4pt}
  \centering
  \caption{Robustness evaluation under different random single-frame annotations. Annotation \#1 is used in all other experiments.}
    \begin{tabular*}{0.6\hsize}{@{}@{\extracolsep{\fill}}l|ccc|ccc@{}}
    \toprule
    &\multicolumn{3}{c|}{SAMM-LV}&\multicolumn{3}{c}{CAS(ME)$^2$}\\
    Strategy&MaE&ME&Overall&MaE&ME&Overall\\
    \midrule
    Annotation \#1&0.4176&0.2417&0.3705&0.4339&0.1370&0.4023\\
    Annotation \#2&0.4290&0.2122&0.3713&0.4401&0.1622&0.4104\\
    Annotation \#3&0.4104&0.2212&0.3638&0.4243&0.1538&0.3982\\
    \bottomrule
    \end{tabular*}%
  \label{randomlabeleval}%
\end{table*}%

\begin{figure*}[t]
\centering
\includegraphics[width=1.0\textwidth]{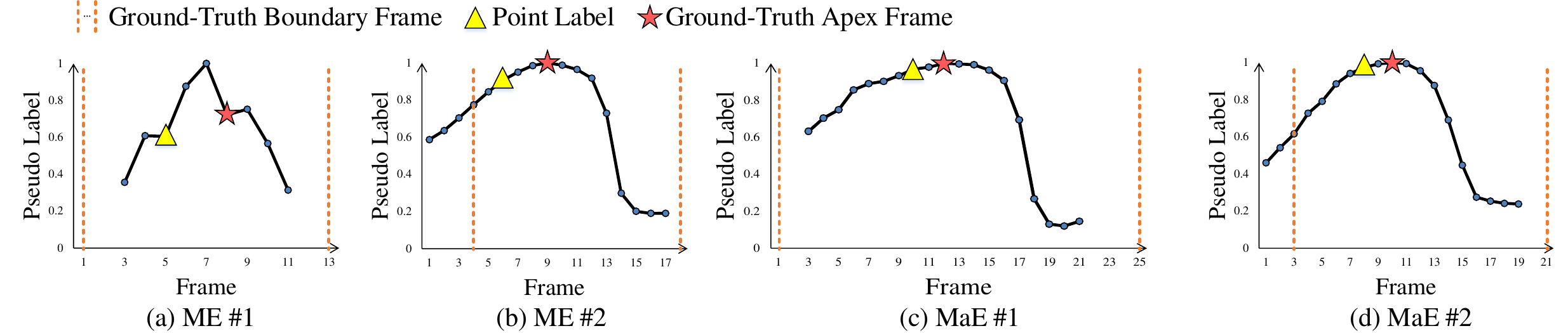}
\caption{Pseudo-label results of four expression instances. The line graph with blue dots represents the soft pseudo-labels assigned by our model; the leftmost and rightmost blue dots indicate the estimated expression duration for pseudo-labeling, while the peak dot indicates the pseudo-apex frame.}
\label{label}
\end{figure*}

\begin{figure*}[t]
\centering
\includegraphics[width=1.0\textwidth]{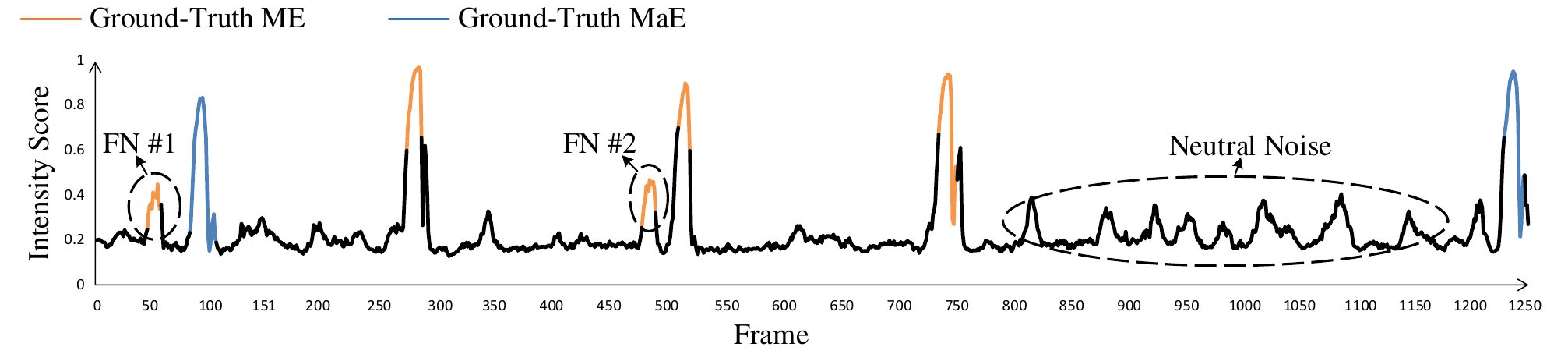}
\caption{Expression intensity results of an entire video.}
\label{intresult}
\end{figure*}

\subsection{Robustness to random single-frame annotation}

In Section~\ref{trainingdetails}, we describe that the single-frame annotations are generated using a Gaussian distribution centered on the ground-truth apex frame for each instance. This process is designed to better simulate real-world annotation scenarios, where even expert annotators may not always label the exact apex frame but rather a nearby frame that appears most representative. The Gaussian distribution naturally models this uncertainty by assigning higher probabilities to frames closer to the true apex.

To evaluate the robustness of our model to such annotation randomness, we generate three independent sets of randomly sampled single-frame annotations and conduct experiments using each set. As shown in Table~\ref{randomlabeleval}, our model achieves consistent performance across these settings, demonstrating the robustness of our proposed framework.

Note that Annotation \#1 in Table~\ref{randomlabeleval} is used consistently across all other experiments and ablation studies. Therefore, the randomness introduced during the annotation generation process does not affect the reproducibility or fairness of our reported results.

\subsection{Qualitative evaluation}
\label{sec:qualitative}
\textbf{Pseudo-label results.}
To provide an intuitive illustration, we present qualitative results of pseudo-labels in Fig.~\ref{label}. The results demonstrate that the pseudo-apex frames detected by the GIM are more close to ground-truth than using point labels. Furthermore, GIM can precisely estimate the duration of expression instances, enabling the assignment of reliable soft pseudo-labels.

\begin{figure}[t]
\centering
\includegraphics[width=0.5\textwidth]{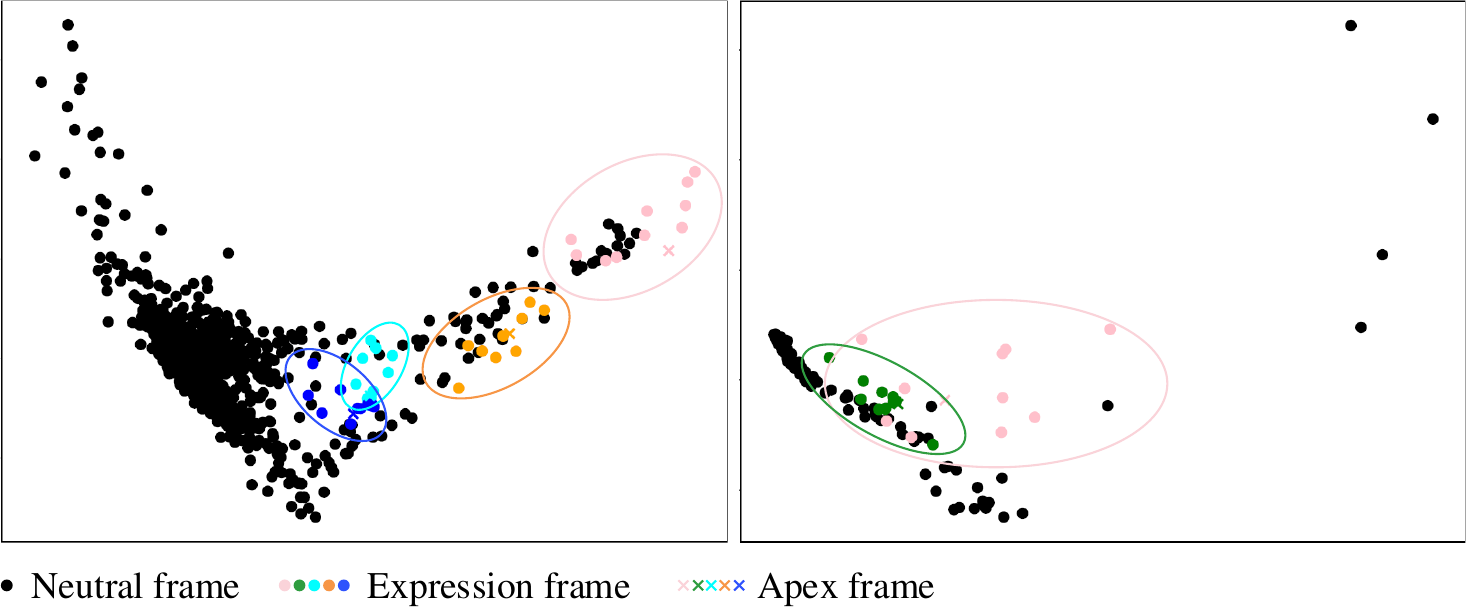}
\caption{PCA distribution of two video examples. Each expression instance is represented by a single color, with the apex frame denoted by $\times$.}
\label{PCA}
\end{figure}

\noindent\textbf{Intensity score results of an entire video.}
As shown in Fig.~\ref{intresult}, we present the expression intensity score results for an entire video. The results indicate that our method effectively suppresses neutral noise while preserving the intensity of expressions, demonstrating its capability to distinguish between expression and neutral frames in most cases. However, some MEs with extremely low intensity may still be difficult to detect, as their intensity scores are weak and nearly indistinguishable from neutral motion fluctuations. These cases illustrate that distinguishing extremely subtle MEs from neutral noise remains an inherent challenge in ME spotting.

\noindent\textbf{Feature space analysis on an entire video.}
We also present qualitative results in the feature space for two example videos, as shown in Fig.~\ref{PCA}. We use Principal Component Analysis (PCA) to examine the feature distribution of all frames in an entire video. Each individual color represents a ground-truth expression instance, while black represents neutral frames. It can be observed that each expression instance follows a Gaussian distribution, with the pseudo-apex frame positioned near the center, demonstrating the effectiveness of our proposed GIM.

\section{Conclusion}
In this paper, we investigated point-supervised facial expression spotting (P-FES). For this purpose, we proposed a two-branch framework that consisted of a regression-based class-agnostic expression intensity branch, which models the expression intensity distribution of each expression instance, and a class-aware apex classification branch, which distinguishes between MaEs and MEs based on pseudo-apex frames. In the expression intensity branch, we introduced a Gaussian-based instance-adaptive Intensity Modeling (GIM) module for soft pseudo-labeling. During training, we detected the pseudo-apex frame around each labeled frame and estimated the rough duration of each expression instance. Then, we built the Gaussian distribution centered at the pseudo-apex frame and assigned soft pseudo-labels to all potential expression frames in the estimated duration. In addition, we introduced an Intensity-Aware Contrastive (IAC) loss on pseudo-neutral frames and pseudo-expression frames with various intensities to enhance feature learning and further suppress neutral noise. Extensive quantitative and qualitative experiments on the SAMM-LV, CAS(ME)$^2$, and CAS(ME)$^3$ datasets demonstrated the effectiveness of our proposed method.


%



\ifCLASSOPTIONcompsoc
  \section*{Acknowledgments}
\else
  \section*{Acknowledgment}
\fi

This work was partially supported by Innovation Platform for Society 5.0 from Japan Ministry of Education, Culture, Sports, Science and Technology, and JSPS KAKENHI Grant Number JP24K03010. This work was also supported by JST SPRING, Grant Number JPMJSP2138.

\ifCLASSOPTIONcaptionsoff
  \newpage
\fi



\bibliographystyle{IEEEtran}
\bibliography{main}
%



%

\begin{IEEEbiography}[{\includegraphics[width=1in,height=1.25in,clip,keepaspectratio]{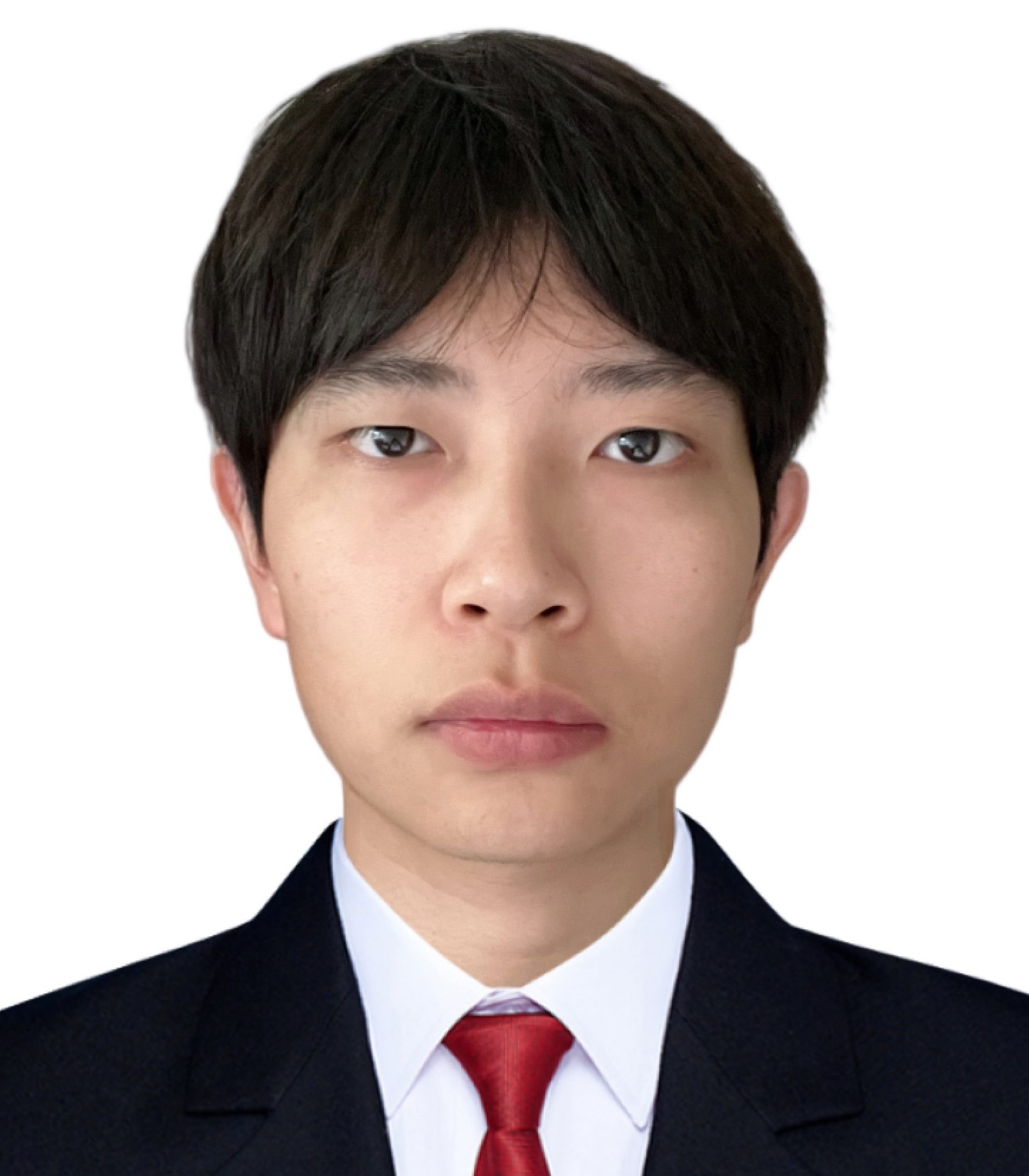}}]{Yicheng Deng} received the B.Eng. and M.Eng. degrees from Beijing Jiaotong University, Beijing, China, in 2019 and 2022, respectively. He is now pursuing his Ph.D. degree in the University of Osaka. His research interests include facial expression analysis and video understanding.\end{IEEEbiography}

\begin{IEEEbiography}[{\includegraphics[width=1in,height=1.25in,clip,keepaspectratio]{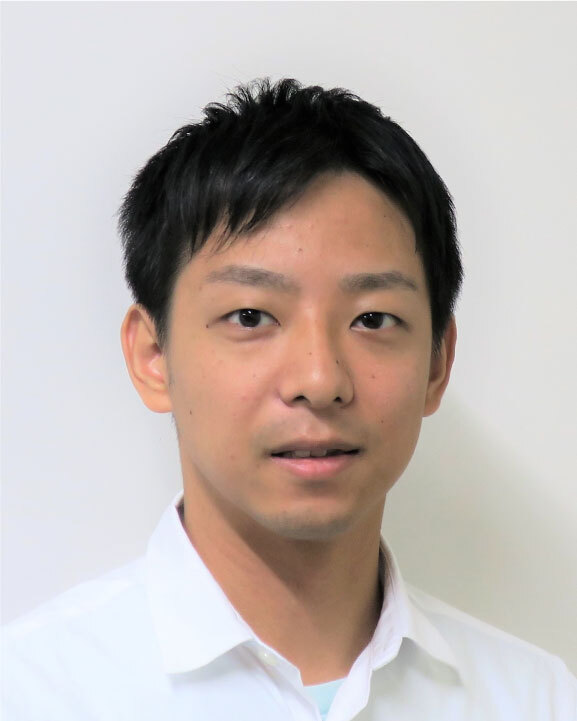}}]{Hideaki Hayashi}
(S' 13--M' 16) received the B.E., M.Eng, and D.Eng. degrees from Hiroshima University, Hiroshima, Japan, in 2012, 2014, and 2016 respectively. He was a Research Fellow of the Japan Society for the Promotion of Science from 2015 to 2017 and an assistant professor in Department of Advanced Information Technology, Kyushu University from 2017 to 2022. He is currently an associate professor with the D3 Center, the University of Osaka. His research interests focus on neural networks, machine learning, and medical data analysis. 
\end{IEEEbiography}


\begin{IEEEbiography}[{\includegraphics[width=1in,height=1.25in,clip,keepaspectratio]{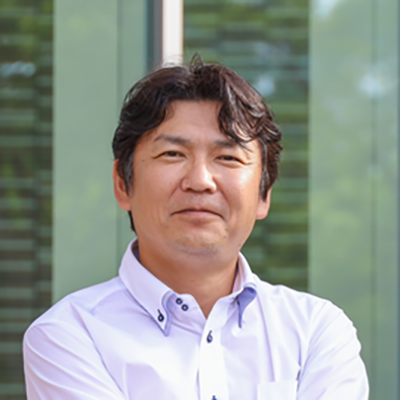}}]{Hajime Nagahara} is a professor at the D3 Center, the University of Osaka, since 2017.  He received Ph.D. degree in system engineering from Osaka University in 2001. He was a research associate of the Japan Society for the Promotion of Science from 2001 to 2003. He was an assistant professor at the Graduate School of Engineering Science, Osaka University, Japan from 2003 to 2010.  He was an associate professor in Faculty of information science and electrical engineering at Kyushu University from 2010 to 2017.
He was a visiting associate professor at CREA University of Picardie Jules Verns, France, in 2005. He was a visiting researcher at Columbia University in 2007-2008 and 2016-2017.
Computational photography and computer vision are his research areas. He received an ACM VRST2003 Honorable Mention Award in 2003, IPSJ Nagao Special Researcher Award in 2012, ICCP2016 Best Paper Runners-up, and SSII Takagi Award in 2016. He is a program chair in ICCP2019, Associate Editor for IEEE Transaction on Computational Imaging in 2019-2022, and Director of Information Processing Society of Japan in 2022-2024.
\end{IEEEbiography}




\end{document}